\documentclass{article}

\usepackage{arxiv}

\usepackage[utf8]{inputenc} 
\usepackage[T1]{fontenc}    
\usepackage{hyperref}       
\usepackage{url}            
\usepackage{booktabs}       
\usepackage{amsfonts}       
\usepackage{nicefrac}       
\usepackage{microtype}      
\usepackage{graphicx}
\usepackage{natbib}
\usepackage{doi}
\usepackage{xcolor}

\usepackage{glossaries}

\newacronym{ai}{AI}{Artificial Intelligence}
\newacronym{cad}{CAD}{Computer-Aided Design}
\newacronym{cade}{CAD/E}{Computer Aided Design and Engineering}
\newacronym{llm}{LLM}{Large Language Model}
\newacronym{mas}{MAS}{Multi Agent System}
\newacronym{vlm}{VLM}{Vision Language Model}

\usepackage{dirtytalk}
\usepackage{color}
\usepackage{listings}
\usepackage{url}
\usepackage[capitalize]{cleveref}
\Crefname{lstlisting}{Listing}{Listings}
\Crefname{algocf}{Algorithm}{Algorithms}

\usepackage{circledtext}
\usepackage[ruled]{algorithm2e}
\SetKwComment{Comment}{/* }{ */}

\usepackage{tabularx}
\usepackage{booktabs}
\usepackage{array}
\usepackage{makecell}
\usepackage{multirow}

\newcommand{\asimo}{\includegraphics[scale=0.08]{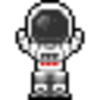}}

\title{From Idea to CAD: A Language Model-Driven Multi-Agent System for Collaborative Design}

\author{
Felix Ocker\textsuperscript{\asimo}
\! Stefan Menzel\textsuperscript{\asimo}
\! Ahmed Sadik\textsuperscript{\asimo}
\! Thiago Rios\textsuperscript{\asimo}
\\
	\asimo Honda Research Institute Europe \\
        Germany, Offenbach am Main, 63073 \\
	\texttt{firstname.lastname@honda-ri.de} \\
}

\renewcommand{\shorttitle}{\textsc{From Idea to CAD}}

\hypersetup{
pdftitle={ocker2025-cadbot},
pdfsubject={cs.AI, cs.MA},
pdfauthor={Felix Ocker},
pdfkeywords={Engineering, Computer Aided Design, Vision Language Model, Multi Agent System},
}

\begin{document}
\maketitle
\begin{abstract}
Creating digital models using Computer Aided Design (CAD) is a process that requires in-depth expertise.
In industrial product development, this process typically involves entire teams of engineers, spanning requirements engineering, CAD itself, and quality assurance.
We present an approach that mirrors this team structure with a Vision Language Model (VLM)-based Multi Agent System, with access to parametric CAD tooling and tool documentation.
Combining agents for requirements engineering, CAD engineering, and vision-based quality assurance, a model is generated automatically from sketches and/ or textual descriptions.
The resulting model can be refined collaboratively in an iterative validation loop with the user.
Our approach has the potential to increase the effectiveness of design processes, both for industry experts and for hobbyists who create models for 3D printing.
We demonstrate the potential of the architecture at the example of various design tasks and provide several ablations that show the benefits of the architecture's individual components.
\end{abstract}


\section{Introduction}

In modern product development, \gls{cade} plays a key role to turn innovative ideas and visions into tangible and manufacturable designs.
Digital 2D and 3D geometry representations of objects on different levels of granularity are required in various intermediate development steps, for example aesthetic discussions, design quality evaluations based on simulation tools, and design feasibility checks.
For these steps, development teams include various roles such as requirement engineers, style designers, \gls{cad} experts, simulation domain experts, and quality assurance teams who create a product cooperatively.
Stakeholders in these roles utilize software tools to implement digital representations of products, also referred to as digital twins.
This process receives an increasing amount of support in the form of \gls{ai} methods.
For example, data science methods provide efficient ways to improve the problem understanding, e.g., by calculating design sensitivities towards a certain performance aspect \citep{PUBA261}, or displaying the distribution of design variations in the solution space using clustering \citep{pub4326pub4374}.
Further machine learning techniques, such as neural networks or regression models, are commonly used to predict design behavior based on simulation data and coupled with computation optimization algorithms to optimize solutions or enhance decision making \citep{10.1145/3319619.3326890}.

With impressive advances in \glspl{llm}, traditional machine learning approaches in \gls{cade} can be extended and complemented with a new level of user interaction and collaboration.
Knowledge democratization and intuitive user interfaces play a pivotal role to open up tools and processes formerly operated by highly-skilled domain experts to a broader user community. 
\glspl{llm} can translate non-technical language into software for downstream applications and explain technical results to users in a level of detail adjusted to their knowledge and skill set.
This paper proposes a human-AI team framework to collaboratively create CAD models based on a sketch and/~or text input.
We realize parts of the team as a \gls{mas}, which consists of several \gls{llm}-based agents interpreting input sketches and images as well as textual descriptions, making plans for \gls{cad} modeling and generating \gls{cad} code, and comparing the resulting \gls{cad} models with given design requirements. 
The agents are implemented so that they can identify missing information and interact with the user through a chat interface to fill gaps, but also exchange information among each improve the design and increase the quality of the \gls{cad} model.

With this paper, we make two contributions in the field of \gls{llm}-based support for engineering.
First, we present a \gls{vlm}-based \gls{mas} for \gls{cad} model generation that mirrors key phases in established human development processes, and show its benefits over a more naive invocation of \glspl{llm}.
Second, we introduce a way to cope with the limitations of \glspl{vlm} regarding spatial reasoning in the context of generating 3D models via visual self-feedback and human validation.
Our architecture is applicable to a wide variety of scenarios, ranging from engineers quickly generating models from sketches drawn in interactive design sessions to novice users bypassing the high barrier of entry to \gls{cad} and creating models they can realize with consumer-grade 3D-printing equipment.

\section{Related Work}


\subsection{Design processes, CAD/E \& its Automation}

\begin{figure}[tb]
    \centering
    \includegraphics[width=.5\columnwidth]{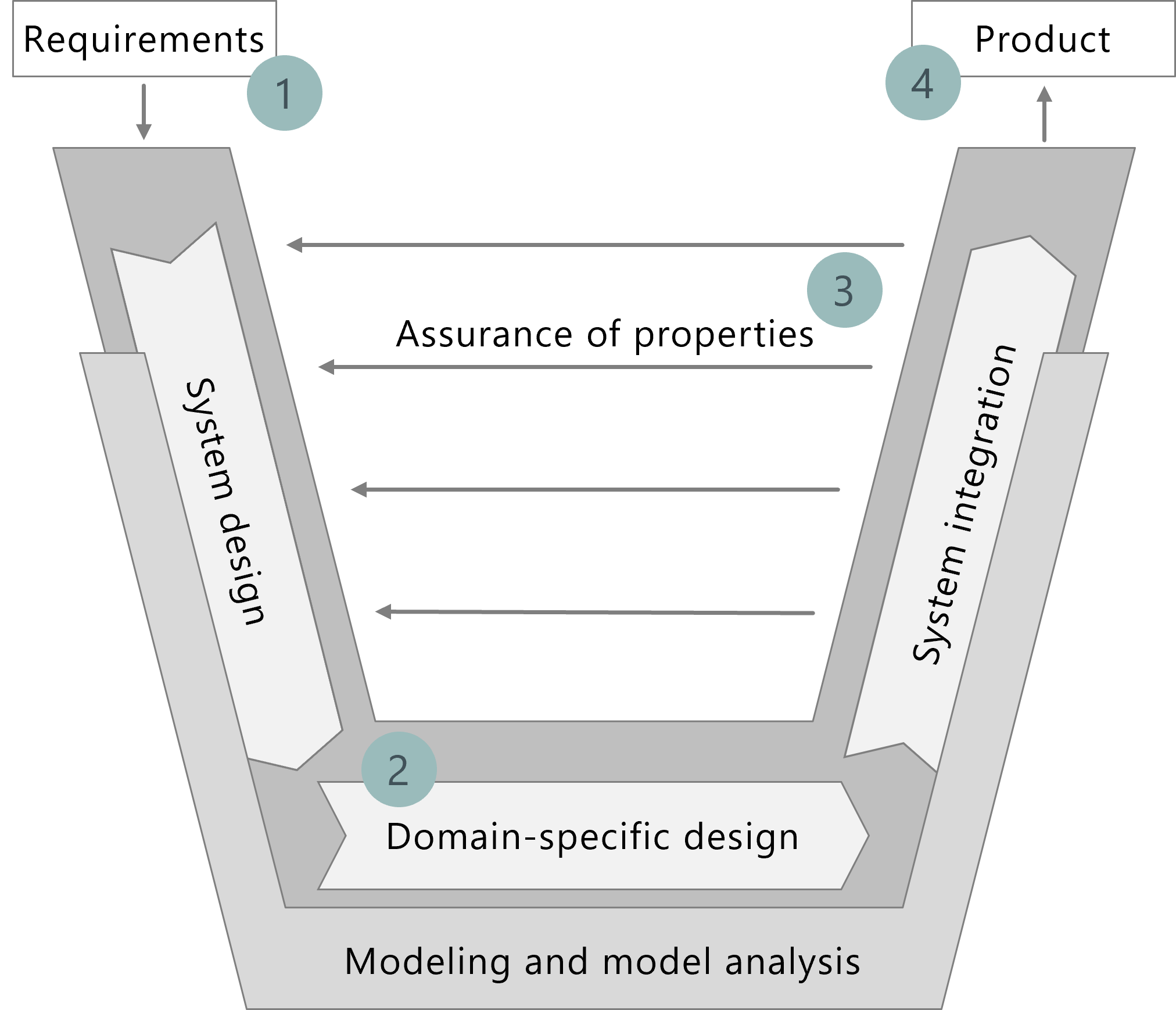}
    \caption{The development phases our approach focuses on highlighted at the example of the V-model: {\circledtextset{resize=real}\circledtext*[height=9pt]{1}}~requirement elicitation, {\circledtextset{resize=real}\circledtext*[height=9pt]{2}}~model creation, {\circledtextset{resize=real}\circledtext*[height=9pt]{3}}~verification, and {\circledtextset{resize=real}\circledtext*[height=9pt]{4}}~validation.}
    \label{fig:vmodel}
\end{figure}

To coordinate the development of complex systems, design processes have been established for engineering.
These range from the V-model \citep{vdi2206} to the waterfall model to more agile ones such as rapid prototyping.
Such development processes share distinct phases for clarifying requirements, creating the system design, iterative requirement compliance checking and design adaptation, and finally handing over the resulting product, cp.~\Cref{fig:vmodel}.
To cope with the complexity of modern engineering systems, computer-aided methods such as \gls{cad} have been developed (e.g., CATIA, AutoCAD, and NX), that enabled the digitization of the design process.
While extremely powerful, \gls{cad} programs typically rely on highly-skilled engineers operating them, may use proprietary data formats, and may incur significant licensing costs.
Alternatively, parametric and code-based part and assembly design, e.g., with the open-source Python library \texttt{CadQuery} \citep{cadquery2024}, allows for the automation and customization of many design processes, while still enabling the use of the generated models in downstream design tasks, such as computer-based physics simulations.

\subsection{Language Models in Engineering}

\glspl{llm} such as GPT \citep{brown2020language} and Llama \citep{dubey2024llama} are advanced \gls{ai} models based on deep learning architectures trained on vast amounts of text data to understand, generate, and analyze human language with context awareness.
Due to their natural language interfaces, \glspl{llm} are intuitive to use by humans and allow for a broad range of applications such as chat bots. 
In addition, \glspl{llm} can be fine-tuned for specific tasks to extend their applicability.
\glspl{vlm} are multi-modal models which are pretrained with large-scale image-text pairs to solve a variety of vision tasks, such as image classification, object detection, and semantic segmentation \citep{zhang2024visionlanguagemodelsvisiontasks}.

Similar to other domains, design and engineering start to integrate \glspl{llm} and other generative \gls{ai} models into existing processes and evaluate their capabilities.
Text-to-image models such as Stable Diffusion and Dall-E \citep{zhang2024texttoimagediffusionmodelsgenerative} support the ideation process and exploration of design variations.
More recently, text-to-3D models have been introduced, enabling the generation of 3D geometries from text prompts.
Shap-E\footnote{\url{https://github.com/openai/shap-e}} extends Point-E to generate surface meshes instead of point clouds.
BlenderGPT \citep{grass2024blendergpt} is a web-based application, which allows the user to generate 3D models with textures based on text prompts or images, and the resulting models can be directly utilized in Blender.
Shape-E has been utilized in combination with \glspl{vlm} for prompt evolution in engineering design optimization \citep{pub-5593}, and \glspl{llm} solely as recombination and mutation operators in multi-task evolutionary design optimization \citep{pub-5773}.


While the above text-to-3D models are based on models pre-trained on 3D objects and result in non-editable models, alternative approaches rely on using parametric \gls{cad} code to generate 3D geometries.
Picard et al. \citep{picard2023concept} assessed \glspl{vlm} on several engineering tasks, including \gls{cad} generation.
Starting from a technical drawing, a \gls{vlm} was prompted to generate code.
They checked the syntax and set up a visual feedback loop with four views of the generated model.
However, the experiment was limited to a single drawing of a block with a single hole.
The authors found that visual feedback did not help and that \glspl{vlm} performed poorly at generating \gls{cad} code in their experiment.
%
%
Alrashedy et al. \citep{alrashedy2024generating} proposed a framework for generating 3D objects with iterative improvements using a \gls{vlm} to generate Python code for parametric \gls{cad}.
The prompt is improved over time through feedback from a \gls{vlm} regarding the quality of the generated 3D object.
Note that the initial prompt contains a prescription of how the code should be created, as well as few-shot demonstrations for guidance, reducing the system's flexibility.
A similar approach is LLM4CAD \citep{li2024llm4cad}, which takes drawings and text as inputs and uses a debugger for the generated code.
This approach relies on OpenAI models and \texttt{CadQuery} and focuses on five types of mechanical components such as shafts, gears, and springs.
The system generated \gls{cad} models that matched the target components, apart from the designs of gears and springs, in which the system lacks performance.
Note that this approach does not include iterations for improving the model created and assumes an initially complete specification.



In contrast to approaches that leverage vanilla \glspl{llm} and \glspl{vlm}, there are also efforts to train models specifically for generating editable \gls{cad} models.
For instance, CAD-LLM \citep{wu2023cad} is a fine-tuned \gls{llm} for generating 2D sketches.
Yuan et al. \citep{yuan2024openecad} created a dataset for sketches and parametric \gls{cad} code and fine-tuned a \gls{vlm} for generating pythonocc\footnote{\url{https://github.com/tpaviot/pythonocc-core}} code.
However, such efforts focus on the \glspl{vlm} themselves, without complementary architecture such as feedback loops.


\begin{figure*}[t]
   \centering
   \includegraphics[width=\textwidth]{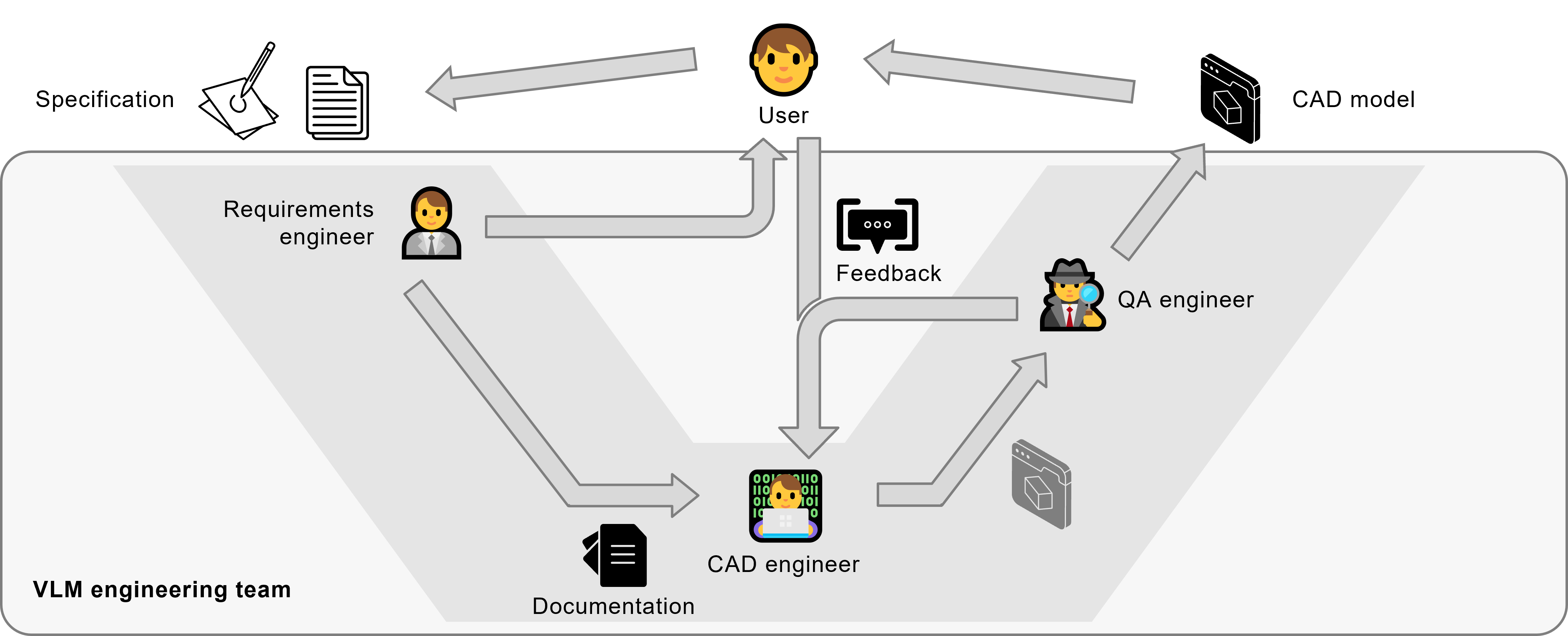}
   \caption{Engineering team architecture.}
   \label{fig:architecture}
\end{figure*}

\subsection{Agents \& Multi Agent Systems}

Intelligent agents are characterized by their autonomy, encapsulation, and goals \citep{wooldridge1995intelligent}.
With \glspl{llm} being available more broadly, there has been a shift from agents realized, e.g., using state-machines, to agents built on \glspl{llm} \citep{ng2024agentic}.
One design pattern, that has been shown to be especially valuable for reasoning with \glspl{llm} is reflexion \citep{shinn2024reflexion}, i.e., the \gls{llm} considers internal and external feedback for correction.
This idea has been demonstrated across various application domains, including robotics \citep{joublin2024copal}.
Multiple such intelligent agents that cooperate to solve complex problems in dynamic and distributed environments are referred to as \glspl{mas}.
Applications of \glspl{mas} range from process automation to engineering automation, manufacturing, and energy systems design.
These applications have in common that decentralized decision-making and cooperation are key.
In engineering, an early framework of 3D \gls{cad} environments enriched by a \gls{mas} has been proposed by Yabuki et al. \citep{2003299}, who link different agents to a \gls{cad} system for steel frame member design. 
While the user develops design proposals, the agents check compliance with the design code and external constraints.
In case of violations, they communicate with the user through the \gls{cad} GUI to improve design aspects.
The developments in \gls{llm}-based agents have also paved the way for \gls{llm}-based \gls{mas}.
Park et al. \citep{park2023generative} were among the first to mimic groups of humans in a simulation with \gls{llm}-based agents.
This paradigm has later been applied to software engineering \citep{qian2024chatdev}, with the agents mimicking teams of specialized humans working together in a software company.

\subsection{Research Gap}

This work builds on advanced technologies in the domains of \gls{cad} and \glspl{llm}.
However, to the best of our knowledge, \gls{cad} models have not yet been created by autonomous \gls{llm}-based \gls{mas}, explicitly mimicking key steps in established human development processes, such as the V-model \citep{vdi2206}.

\section{VLM-Based MAS for CAD}

\subsection{Architecture Overview}

To partially automate the development process using \gls{cad}, we propose a collaborative system that goes through the core phases in established development processes, cp.~\Cref{fig:vmodel}.
Specifically the requirements elicitation {\circledtextset{resize=real}\circledtext*[height=9pt]{1}}, the actual design~{\circledtextset{resize=real}\circledtext*[height=9pt]{2}}, the verification of the model created {\circledtextset{resize=real}\circledtext*[height=9pt]{3}}, and the validation with the user {\circledtextset{resize=real}\circledtext*[height=9pt]{4}}.
The system combines three specialized agents to represent the different roles throughout the development process, cp.~\Cref{fig:architecture}.
The \texttt{RequirementsEngineer} serves as an interface between the user and the more technical \texttt{CadEngineer}.
It clarifies the specification and resolves ambiguities interactively with the user.
The \texttt{CadEngineer} is responsible for creating the \gls{cad} model.
Finally, the \texttt{QualityAssuranceEngineer} verifies the results and provides feedback to the \texttt{CadEngineer}.
Eventually, the model is passed to the user, who may request further changes in an outer validation loop.

\subsection{Requirement Specification}
\label{subsec:requirementspec}

The design process starts with a specification provided by the user.
This initial specification may be a sketch, a purely textual description of the part, or a combination of both.
During the requirements specification phase, the \texttt{RequirementsEngineer} cooperates with the human to identify and resolve all ambiguities in the initial specification iteratively.
The agent checks visual and textual inputs, and asks for further information until the model is sufficiently specified with regards to geometric features such as shapes, orientations, and dimensions.
Note that humans may rely on the \gls{vlm} to make reasonable assumptions.
This reduces the effort for the human but may lead to greater deviations from the intended result.
\Cref{alg:reqs} summarizes the process of specifying the requirements and $prompt_{clarify}$ is available in \Cref{lst:rq}.

\begin{algorithm}
\caption{Interactive requirement specification.}\label{alg:reqs}
\KwData{$sketch \ \mathcal{S}, text \ \mathcal{T}$}
\KwResult{$specification \ \mathcal{R}$}
$ambiguities \gets vlm.prompt(prompt_{clarify}, \mathcal{S}, \mathcal{T})$\;
\While{$ambiguities \neq \emptyset$}{
  $output(ambiguities)$\;
  $\mathcal{T} \gets \mathcal{T} + user.input()$\;
  $ambiguities \gets vlm.prompt(prompt_{clarify}, \mathcal{S}, \mathcal{T})$
}
$\mathcal{R} \gets (\mathcal{S}, \mathcal{T})$\;
\end{algorithm}

\subsection{Model Design}

Upon clarification of the requirements, the specification is passed on to the \texttt{CadEngineer}.
Based on the specification, the \texttt{CadEngineer} comes up with a coarse textual plan for the modeling process.
Following this plan, the \texttt{CadEngineer} then generates Python code using the \gls{cad} library \texttt{CadQuery}.
The generated code is checked for executability and is eventually executed.
In both cases, the \gls{vlm} is prompted to revise the code with regards to the error messages, if necessary.
Note that the code execution can be set to require human confirmation to avoid potentially security-critical code execution.
This process results in a \texttt{.stl} file, which is saved on disk.
Note that the \texttt{CadEngineer} is also able to take into account feedback from the verification phase, i.e., from the \texttt{QualityAssuranceEngineer}, and the validation phase, i.e., from the user.
In case such feedback has been provided, the \texttt{CadEngineer} checks the documentation for potential misuse of the package and makes use of hints for regenerating the code.
\Cref{alg:cad} summarizes the process of iteratively designing the model and the prompts $prompt_{plan}$ and $prompt_{code}$ are available in \Cref{lst:plan,lst:cad}, respectively.

\begin{algorithm}
\caption{Model design.}\label{alg:cad}
\KwData{$specification \ \mathcal{R}, docs \ url_{docs}, feedback \ \mathcal{F}$}
\KwResult{$code \ \mathcal{C}, model \ \mathcal{M}$}
$docs \gets retrieve(url_{docs})$\;
$hints \gets \emptyset$\;
\If{$\neg \mathcal{F}$}
{
  $plan \gets vlm.prompt(prompt_{plan}, \mathcal{R})$\;
}
\While{$\neg \mathcal{M}$}{
  \If{$\mathcal{F}$}
  {
    $hints \gets llm.prompt(prompt_{docs}, docs, \mathcal{F})$\;
  }
  $\mathcal{C} \gets vlm.prompt(prompt_{code}, \mathcal{R}, hints)$\;
  \If{$check(\mathcal{C})$}
  {
    $\mathcal{M} \gets exec(\mathcal{C})$\;
  }
}
\end{algorithm}

\subsection{Verification}

To cope with limitations of \glspl{vlm} with regards to spatial reasoning and help in the process of translating the specification into code, we make use of a verification loop.
This verification loop equates to the verification phase in human development cycles, where the model is compared to the initial specification to ensure that the design fulfills all requirements.
Accordingly, the \texttt{QualityAssuranceEngineer} takes as an input the model created by the \texttt{CadEngineer} and creates various views of the model.
The default is a set of seven images, taken from the top, bottom, right, left, front, back, and an isometric one.
The \texttt{QualityAssuranceEngineer} then compares these views of the model with the specification and checks if the specification is met.
In case there are deviations from the specification, the \texttt{QualityAssuranceEngineer} provides a list of textual suggestions on what should be changed.
\Cref{alg:ver} summarizes the verification loop and the prompt $prompt_{qa}$ is available in \Cref{lst:qa}.

\begin{algorithm}
\caption{\gls{vlm}-based verification.}\label{alg:ver}
\KwData{$specification \ \mathcal{R}, user \ feedback \ \mathcal{F}_{val}$}
\KwResult{$feedback_{verification} \ \mathcal{F}_{ver}, verified \ model \ \mathcal{M}$}

$\mathcal{F}_{ver} \gets \emptyset$\;

\While{$\top$}{
  $\mathcal{M} \gets design(\mathcal{R}, \mathcal{F}_{val} + \mathcal{F}_{ver})$ \Comment*[r]{\Cref{alg:cad}}
  $views \gets render(model)$\;
  $\mathcal{F}_{ver} \gets vlm.prompt(prompt_{qa}, \mathcal{R}, views)$\;
  \If{$\mathcal{F}_{ver} = \emptyset$}{
    $break$\;
  }
}

\end{algorithm}

\subsection{Validation}

Despite the initial interactive requirements specification, cp.~\Cref{subsec:requirementspec}, the specification may be incorrect or the system may be unable to converge towards the desired model.
To cope, we introduce an outer validation loop, analogous to the validation in development processes \citep{vdi2206}.
Here, the user is asked to confirm the model created, and asked for specific feedback in case they are not satisfied.
This human feedback is used to regenerate the model, cp.~\Cref{alg:val}.

\begin{algorithm}
\caption{Human validation.}\label{alg:val}
\KwData{$specification \ \mathcal{R}$}
\KwResult{$validated \ model \ \mathcal{M}$}

$\mathcal{F}_{val} \gets \emptyset$\;

\While{$\top$}{
  $\mathcal{M} \gets design(\mathcal{R}, \mathcal{F}_{val})$ \Comment*[r]{\Cref{alg:ver}}
  $output(\mathcal{M})$\;
  $\mathcal{F}_{val} \gets user.input()$\;
  \If{$\mathcal{F}_{val} = \emptyset$}{
    $break$\;
  }
}

\end{algorithm}

\begin{figure*}[t]
    \centering
    \includegraphics[width=\textwidth]{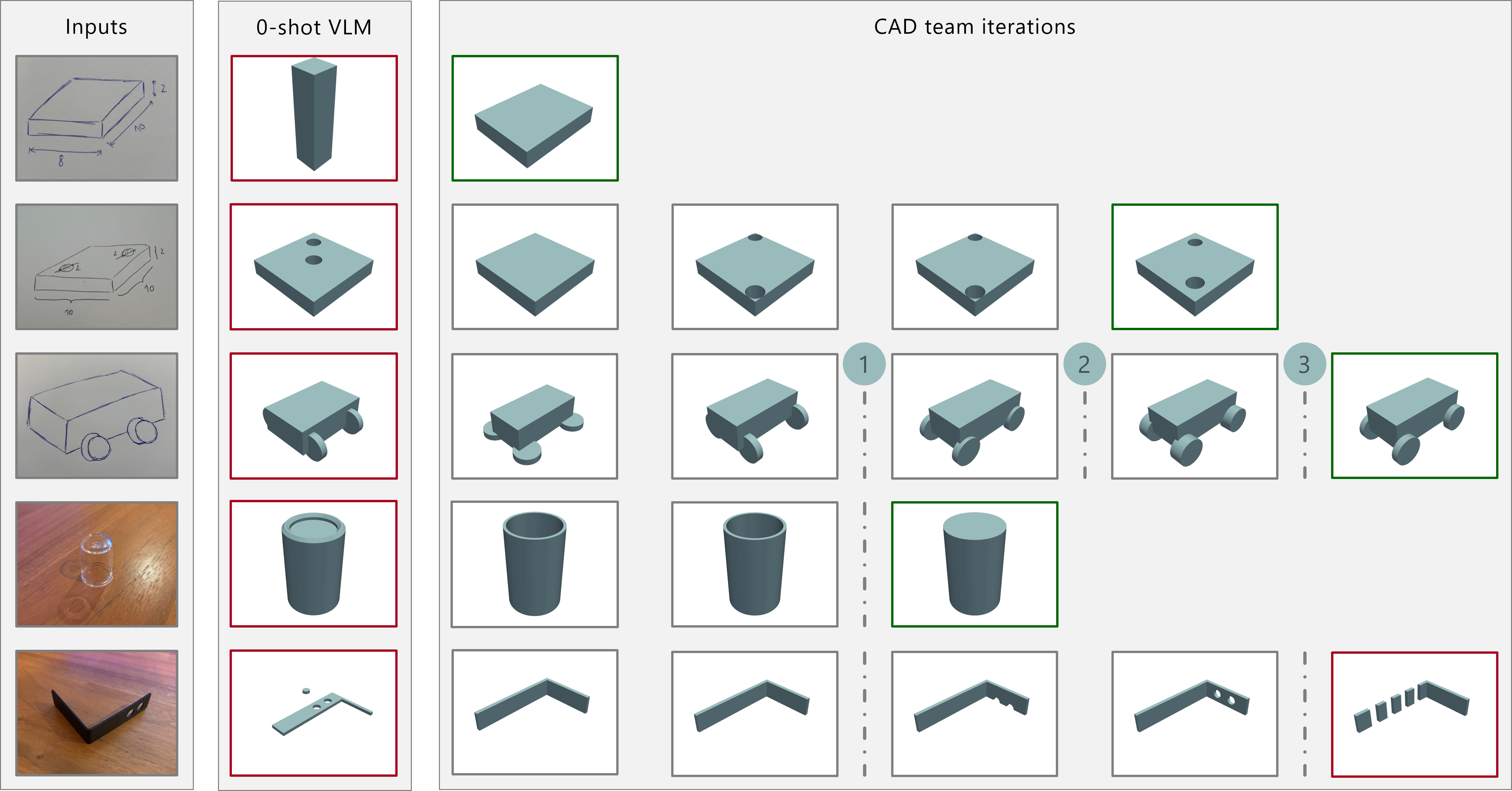}
    \caption{Examples of the iterative design process from the user specification to the resulting model. The first column \textit{``Inputs''} shows the visual inputs, the second column \textit{``0-shot \gls{vlm}''} shows the baseline results without requirement clarification, verification and validation, and the right part shows the iterations with our \gls{mas}. The dashed lines indicate validation steps. Red borders indicate wrong \gls{cad} models, green borders indicate acceptable designs.}
    \label{fig:examples}
\end{figure*}

\section{Implementation, Experiments \& Discussion}

\subsection{Implementation}

We implemented a prototype in Python, relying on several well-established packages.
For code-based parametric model design, we used the Python library \texttt{CadQuery} \citep{cadquery2024}, which comes with detailed documentation\footnote{\url{https://cadquery.readthedocs.io/en/latest/classreference.html\#classreference}} that can be fed into a \gls{vlm}.
For checking the code generated before executing it, we relied on Python's \texttt{ast} module.
\texttt{CadQuery}'s documentation is retrieved from the web using Python's \texttt{request} package in combination with \texttt{BeautifulSoup4}\footnote{\url{https://www.crummy.com/software/BeautifulSoup/}}.
We render stl files within the \texttt{QualityAssuranceEngineer} leveraging \texttt{pyvista}\footnote{\url{https://pyvista.org/}}.
As \glspl{vlm}, we rely on OpenAI's \texttt{gpt-4o-2024-11-20} as a default, interacting with the API via OpenAI's Python package\footnote{\url{https://github.com/openai/openai-python}}.
Note that error handling for the \gls{vlm} API was crucial for robust execution.

\subsection{Experiments}

We conducted experiments with visual inputs of varying complexity, both for hand-drawn sketches and photos of actual objects.
Here, the number of geometrical features to be realized serves as an indication for complexity.
As hand-drawn sketches, we included a simple rectangular block, a rectangular block with two holes, and a strongly simplified sketch of a toy car.
We also included photos, one of a plastic cap and one of an angle bracket.
\Cref{fig:examples} gives an overview of these inputs on the left, and \Cref{tab:texts} shows the textual parts of the specifications.

\begin{table}
\caption{Complementary textual descriptions for the visual inputs depicted in \Cref{fig:examples}.}
\label{tab:texts}
\begin{center}
    \begin{tabularx}{\linewidth}{ l | X }
    \textbf{Part} & \textbf{Textual description} \\
    \hline
    Block & \textit{-} \\
    \hline
    Block w/ holes & \textit{length=10 width=10 height=2, all in cm. the holes are through-holes with diameter 2cm. they are positioned in opposite corners of the part, 2cm away from the closest edge. the material is plastic.} \\
    \hline
    Simple car &  \textit{a simplified model of a toy car. length=12cm width=8cm including wheels height=6cm including wheels. wheels have a diameter of 4cm and a width of 1cm. make reasonable assumptions for all other dimensions! ignore further details.} \\
    \hline
    Cap & \textit{Make reasonable assumptions for all dimensions and other features. Keep it simple - limit the design only to the absolutely necessary features.} \\
    \hline
    Angle bracket & \textit{short leg length = 3cm, long leg length = 5cm. leg width = 1cm. thickness = .2cm. angle is 90 degrees. the holes have a diameter of 0.5cm and are 1cm apart. the outer hole is 1cm apart from the legs end. material is aluminium. disregard all other details.} \\
    \end{tabularx}
\end{center}
\end{table}

To assess the benefits of the individual agents in the system, we performed ablations using the selected designs.
The second column ``0-shot \gls{vlm}'' shows the performance of the \texttt{CadEngineer} by itself, without interactive requirement specification, verification, and validation.
This setup serves as a baseline, as it relies on the same prompt and utilizes the planning step before the code generation.
The designs in the third column, i.e., the first column of the ``\gls{cad} team iterations'' correspond to the designs created by the \texttt{CadEngineer} together with the \texttt{RequirementsEngineer}, but without any iterations.
All further columns correspond to further iterations.
For an example of the code generated for the part with two holes see~\Cref{lst:code}.
During the validation steps indicated by dashed lines, the user provided further feedback.
For instance, see \Cref{tab:validation} for the feedback given in the design process of the toy car model.

\begin{table}
\caption{User feedback in the validation steps for the toy car model.}
\label{tab:validation}
\begin{center}
    \begin{tabular}{ l | l }
    \textbf{Step} & \textbf{User feedback} \\
    \hline
    1 & \textit{make the wheels parallel to the XZ plane, not the XY plane!} \\
    \hline
    2 &  \textit{the wheels are asymmetric - create them by extruding into \_both\_ directions} \\
    \hline
    3 & \textit{make the wheels only half as wide} \\
    \end{tabular}
\end{center}
\end{table}


\subsection{Discussion}

Overall, the results of the \gls{vlm}-based \gls{mas} are promising, especially when compared to the single shot generative approach.
By utilizing specialized agents that mimic typical engineering roles and a self-feedback loop in the architecture, our method generates designs with a higher level of readiness and compliance to the requirements.
Nonetheless, there is significant variability in the results and making such \gls{vlm}-based systems with self-feedback robust is still challenging for several reasons:
In line with limitations regarding spatial reasoning, \glspl{vlm} still seem to struggle especially with the orientation of surfaces, e.g., when selecting suitable workplanes for sketches.
Also, the \gls{mas} failed to generate more complex components.
Note that simply increasing the number of iterations may be insufficient, as this would pollute the context window with faulty examples, thus hindering convergence.
Finally, the overall system performance is also limited by the capabilities of the parametric \gls{cad} libraries used.
For instance, more advanced \gls{cad} tools enable bending operations, which could facilitate the angle bracket design, while \texttt{CadQuery} requires a different approach, since the tool lacks shape-morphing operations.
This became apparent, when the \gls{vlm} hallucinated a \texttt{.bend} operation, which would also be an intuitive design choice for an engineer.

\section{Summary and Outlook}

This paper presented an \gls{llm}-based \gls{mas} architecture for \gls{cad} inspired by established development processes.
The system combines iterative requirement specification with a human user in the loop, code generation from sketches and text for parametric \gls{cad} using the documentation, a verification loop using visual feedback, and a validation loop with the user.
Our experiments showed that the combination of these components enables the generation of designs with a significantly higher readiness level than the vanilla \gls{vlm} for \gls{cad} model generation.

To further improve the system, several research directions still remain to be explored.
First, the code generation might benefit from a more fine-grained iterative process for translating the design plan into actual code, potentially using a visual confirmation loop to create the model's individual features step by step.
This step might also benefit from adding further spatial context for the \gls{vlm}, such as including a coordinate system and possibly even dimensions in every rendering.
Second, for expert users, it may be beneficial to be able to edit the generated code, possibly even in an interactive way.
Finally, vanilla \glspl{vlm} exhibit impressive capabilities, even when generating code for specific packages such as \texttt{CadQuery}.
Nonetheless, the individual agents in the system, and specifically the \texttt{CadEngineer}, are likely to benefit from using a \gls{vlm} finetuned on a dedicated training corpus.


\bibliographystyle{unsrtnat}
\bibliography{references}

\clearpage

\section*{Appendix}

\Cref{lst:rq,lst:cad,lst:plan,lst:docs,lst:qa} show the prompts used for the individual agents and \Cref{lst:code} shows the code generated for the example, cp.~\Cref{fig:examples}.

\begin{lstlisting}[label=lst:rq, language=Python, caption=Requirements agent prompt.]
RE_PROMPT = """\
You are an expert requirements engineer in the preparation phase of doing CAD work.
Given a description of a part, identify all insufficiently specified aspects.
Discuss and resolve all issues together with the user.
Focus on the parts that are relevant for the CAD engineer, i.e., dimensions and positions.
Do not write CAD code just yet, but clarify the specification with the user by asking questions

ONLY when everything is clear, return a summary of the missing information.
Do so by writing an addendum to the specification that summarizes what you learned from the user.
This summary must be enclosed between HTML-like structural elements: <SUMMARY> the summary </SUMMARY>.
You may ONLY use the <SUMMARY> keyword for returning the final addendum to the specification.
"""
\end{lstlisting}

\begin{lstlisting}[label=lst:cad, language=Python, caption=CAD agent prompt.]
CAD_ENGINEER_PROMPT = """\
You are an expert CAD engineer with access to the Python library CadQuery.
Your job is to create Python code that generates a 3D model based on a given description.
The description may be textual or in the form of an image, e.g., hand drawn.
Make sure to include all relevant parts.
Pay special attention to the orientation of all parts, e.g., by choosing appropriate workplanes.
For instance, pick a workplane perpendicular to the ground for sketching the outline of the wheels of a toy car.
Whenever possible, use the default workplanes, i.e., XY, XZ, and YZ.

For instance, for the instruction `Create a block of dimensions 2 x 2 x 2.` \
the code could be:
```Python
import cadquery as cq

length=2
height=2
thickness=2

result = (
    cq.Workplane("XY")
    .box(length, height, thickness)
)
```

Make sure to create the model as `result`.
Return Python code only, no markdown or comments.
"""
\end{lstlisting}

\begin{lstlisting}[label=lst:plan, language=Python, caption=CAD agent planning prompt.]
STRUCTURE_THOUGHTS_PROMPT = """\
You are an expert CAD engineer who is very experienced with the Python library CadQuery.
Given a specification, come up with a plan consisting of the rough steps necessary for creating the model.
Include steps such as the definition of the key planes and sketches, \
as well as the extrusion steps and definitions of parametric curves.
Return a numbered list of these relevant steps.
"""
\end{lstlisting}

\begin{lstlisting}[label=lst:docs, language=Python, caption=CAD agent documentation retrieval prompt.]
DOCUMENTATION_RETRIEVAL_PROMPT = """\
In the following, you are provided with code, feedback, and documentation.
If applicable, make suggestions for fixes to the code using the documentation.

Code:
{code}

Feedback:
{feedback}

Documentation:
{documentation}

Return concrete suggestions of what should be changed in the code.
"""
\end{lstlisting}

\begin{lstlisting}[label=lst:qa, language=Python, caption=QA agent prompt.]
QAE_PROMPT = """\
You are a quality assurance engineer tasked with reviewing a 3D model with regards to the specification.
The specification may be textual or in the form of a sketch.
The model is available in seven views: from the top, bottom, front, back, left side, right side, and isometric.
Compare the model with the specification and identify all relevant discrepancies regarding the geometry, \
such as orientation and adjacency of parts.
Identify the two most relevant issues and provide concrete suggestions for changes to be made, e.g.:
1. the cylinder is orientied incorrectly, it should be turned by 90 degrees
2. the hole is positioned incorrectly, it should be closer to the edge

If the model is acceptable, return an empty string.
"""
\end{lstlisting}

\begin{lstlisting}[label=lst:code, language=Python, caption=Code generated for the example.]
import cadquery as cq

# Step 1: Create the base block (non-centered)
block = cq.Workplane("XY").box(10, 10, 2, centered=False)
# Step 2: Define the hole positions (relative to the block's bottom-left corner)
hole_positions = [(2, 8), (8, 2)]  # (x, y) coordinates of the hole centers
# Step 3: Create the through-holes on the top face
block = block.faces(">Z").workplane().pushPoints(hole_positions).hole(2)
# Step 4: Finalize the model
result = block

result.export("/data/2025-01-14-15-06-38-block-w-holes/example.stl")
\end{lstlisting}

\end{document}